\documentclass[conference]{IEEEtran}
\makeatletter\if@twocolumn\PassOptionsToPackage{switch}{lineno}\else\fi\makeatother

\usepackage{graphicx}
\usepackage{xurl}
\usepackage[T1]{fontenc}
\ifCLASSINFOpdf
\else
\fi
\usepackage{amsmath}
\usepackage[cmintegrals]{newtxmath}
\usepackage{url,multirow,morefloats,floatflt,cancel,tfrupee}
\makeatletter

\AtBeginDocument{\@ifpackageloaded{textcomp}{}{\usepackage{textcomp}}}
\makeatother
\usepackage{colortbl}
\usepackage{xcolor}
\usepackage{pifont}
\usepackage[nointegrals]{wasysym}
\makeatletter

\def\mcWidth#1{\csname TY@F#1\endcsname+\tabcolsep}

\def\cAlignHack{\rightskip\@flushglue\leftskip\@flushglue\parindent\z@\parfillskip\z@skip}
\def\rAlignHack{\rightskip\z@skip\leftskip\@flushglue \parindent\z@\parfillskip\z@skip}

\@ifundefined{etal}{}{}

\usepackage{ifxetex}
\ifxetex\else\if@twocolumn\@ifpackageloaded{stfloats}{}{\usepackage{dblfloatfix}}\fi\fi

\AtBeginDocument{
\expandafter\ifx\csname eqalign\endcsname\relax
\def\eqalign#1{\null\vcenter{\def\\{\cr}\openup\jot\m@th
  \ialign{\strut$\displaystyle{##}$\hfil&$\displaystyle{{}##}$\hfil
      \crcr#1\crcr}}\,}
\fi
}

\AtBeginDocument{%
  \@ifpackageloaded{endfloat}%
   {\renewcommand\efloat@iwrite[1]{\immediate\expandafter\protected@write\csname efloat@post#1\endcsname{}}}{\newif\ifefloat@tables}%
}%

\def\BreakURLText#1{\@tfor\brk@tempa:=#1\do{\brk@tempa\hskip0pt}}
\let\lt=<
\let\gt=>
\def\processVert{\ifmmode|\else\textbar\fi}

\@ifundefined{subparagraph}{
\def\subparagraph{\@startsection{paragraph}{5}{2\parindent}{0ex plus 0.1ex minus 0.1ex}%
{0ex}{\normalfont\small\itshape}}%
}{}

\newcommand\role[1]{\unskip}
\newcommand\aucollab[1]{\unskip}
  
\@ifundefined{tsGraphicsScaleX}{\gdef\tsGraphicsScaleX{1}}{}
\@ifundefined{tsGraphicsScaleY}{\gdef\tsGraphicsScaleY{.9}}{}
\def\checkGraphicsWidth{\ifdim\Gin@nat@width>\linewidth
	\tsGraphicsScaleX\linewidth\else\Gin@nat@width\fi}

\def\checkGraphicsHeight{\ifdim\Gin@nat@height>.9\textheight
	\tsGraphicsScaleY\textheight\else\Gin@nat@height\fi}

\def\fixFloatSize#1{}
\let\ts@includegraphics\includegraphics

\def\inlinegraphic[#1]#2{{\edef\@tempa{#1}\edef\baseline@shift{\ifx\@tempa\@empty0\else#1\fi}\edef\tempZ{\the\numexpr(\numexpr(\baseline@shift*\f@size/100))}\protect\raisebox{\tempZ pt}{\ts@includegraphics{#2}}}}

\AtBeginDocument{\def\includegraphics{\@ifnextchar[{\ts@includegraphics}{\ts@includegraphics[width=\checkGraphicsWidth,height=\checkGraphicsHeight,keepaspectratio]}}}

\DeclareMathAlphabet{\mathpzc}{OT1}{pzc}{m}{it}

\def\URL#1#2{\@ifundefined{href}{#2}{\href{#1}{#2}}}

\def\UrlOrds{\do\*\do\-\do\~\do\'\do\"\do\-}%
\g@addto@macro{\UrlBreaks}{\UrlOrds}

\edef\fntEncoding{\f@encoding}

\makeatother

\newif\ifmultipleabstract\multipleabstractfalse%
\usepackage{tabulary}
\makeatletter
\AtBeginDocument{\@ifpackageloaded{longtable}{%
\def\LT@makecaption#1#2#3{%
  \LT@mcol\LT@cols c{\hbox to\z@{\hss\parbox[t]\LTcapwidth{%
    \sbox\@tempboxa{#1{#2: } #3}%
    \ifdim\wd\@tempboxa>\hsize
      #1{#2: }\textsc{#3}%
    \else
      \hbox to\hsize{\hfil\box\@tempboxa\hfil}%
    \fi
    \endgraf\vskip\baselineskip}%
  \hss}}}
}{}}
\makeatother

\makeatletter
\let\citep\cite
\let\citet\cite
\makeatother

   \makeatletter
  \def\fig@textbf{\textbf}
   \AtBeginDocument{\renewcommand\floatc@plain[2]{\setbox\@tempboxa\hbox{{\footnotesize#1.}\footnotesize\hskip.5em#2}%
    \ifdim\wd\@tempboxa>\hsize {\fig@textbf{\footnotesize#1.}}\footnotesize\hskip.5em#2\par
        \else\hbox to\hsize{\hfil\box\@tempboxa\hfil}\fi}}
    \makeatother

\usepackage{float}

\begin{document}

%


        \title{Credit card score prediction using machine learning models: A new dataset}
      \author{\IEEEauthorblockN{{Anas~Arram}, 
          }\IEEEauthorblockA{\parbox{150pt}{\centering \textit{Department of Computer Science , }\\Birzeit University, Birzeit, Palestine}}\\[-12pt]Email: anas.arram@gmail.com~\\ \and \IEEEauthorblockN{{Masri~Ayob}, 
          }\IEEEauthorblockA{\parbox{150pt}{\centering \textit{Centre for Artificial Intelligence (CAIT), Data Mining and Optimization Research Group (DMO), }\\Universiti Kebangsaan Malaysia, \\Bangi, 43600, Selangor, Malaysia}}\\[-12pt]Email: masri@ukm.edu.my~\\ \and \IEEEauthorblockN{{Musatafa~Abbas Abbood Albadr}, 
          }\IEEEauthorblockA{\parbox{150pt}{\centering \textit{Department of Oil and Gas Engineering, }\\Basrah University for Oil and Gas, Basrah, 61004, Iraq}}~\\ \and \IEEEauthorblockN{{Alaa~Sulaiman}, 
          }\IEEEauthorblockA{\parbox{150pt}{\centering \textit{Pattern Recognition Research Group, Centre for Artificial Intelligence Technology, Faculty of Information Science and Technology, }\\Universiti Kebangsaan Malaysia (UKM), \\Bangi, 43600, Selangor, Malaysia}}\\[-12pt]Email: alaasol@gmail.com~\\ \and \IEEEauthorblockN{{Dheeb Albashish}}\IEEEauthorblockA{\parbox{150pt}{\centering \textit{Computer Science Department, Prince Abdullah bin Ghazi Faculty of Information and Communication Technology, }\\Al-Balqa Applied University, \\Salt, Jordan}}\\[-12pt]Email: bashish@bau.edu.jo~\\ }


\maketitle 

\begin{abstract}
The use of credit cards has recently increased, creating an essential need for credit card assessment methods to minimize potential risks. This study investigates the utilization of machine learning (ML) models for credit card default prediction system. The main goal here is to investigate the best-performing ML model for new proposed credit card scoring dataset. This new dataset includes credit card transaction histories and customer profiles, has been proposed and tested using a variety of machine learning algorithms, including logistic regression, decision trees, random forests, multi layer perceptron (MLP) neural network, XGBoost, and LightGBM. To prepare the data for machine learning models, we employ different data preprocessing techniques such as feature extraction, handling missing values, managing outliers, and applying data balancing methods. Experimental results demonstrate that MLP outperforms logistic regression, decision trees, random forests, LightGBM, and XGBoost in terms of predictive performance in true positive rate, achieving an impressive area under the curve (AUC) of 86.7\% and an accuracy rate of 91.6\%, with a recall rate exceeding 80\%. These results indicate the superiority of MLP in predicting the default customers and assessing the potential risks. Furthermore, they help banks and other financial institutions in predicting loan defaults at an earlier stage.
\end{abstract}
    

%
\IEEEpeerreviewmaketitle

\section{Introduction}
Credit score is a statistical model used by lenders to assess the creditworthiness of potential borrowers. The model typically uses a variety of factors, such as the borrower's credit history, income, and debts, to generate a score that represents the borrower's likelihood of repaying a loan. Lenders use credit scores to make decisions about whether to approve a loan application and, if so, what interest rate to charge. Credit scores can also be used to determine the credit limits on a credit card. There are a number of different credit scoring models, but they all share some common features. First, they all use a combination of positive and negative factors to assess a borrower's creditworthiness. Second, they all assign a higher score to borrowers who have a history of paying their debts on time and in full. Third, they all take into account the borrower's current financial situation, such as their income and debts \unskip~\cite{2069413:29027063,2069413:29032527,2069413:29164640}.

In recent years, machine learning models have been increasingly used in credit scoring systems to help identify and predict potential default customers. These models use a variety of data, such as the customer's credit history, income, and debts, to generate a score that represents the customer's likelihood of defaulting on a loan. Machine learning models can be more accurate than traditional credit scoring models in predicting default risk. This is because machine learning models can take into account a wider range of data and can learn from past data to improve their accuracy \unskip~\cite{2069413:29027098,2069413:29027570,2069413:29027572}.

Many machine learning models have been applied for predicting credit scores and default customers. Bahnsen \textit{et al.}\unskip~\cite{2069413:29028116} proposed an example-dependent cost matrix for credit scoring, which incorporates all the real financial costs associated with the lending business. They employed logistic regression as the prediction model by modifying the model's objective function to be cost-sensitive.  The proposed model was evaluated against two publicly available datasets:  2011 Kaggle competition Give Me Some Credit and  2009 Pacific-Asia Knowledge Discovery and Data Mining conference (PAKDD) competition. Their results showed that their proposed method outperformed the state-of-the-art methods. 

Butaru \textit{et al.}\unskip~\cite{2069413:29030518} conducted a comprehensive study employing various machine learning models to assess credit card risks. These models included decision trees, regularized logistic regression, and random forests. The evaluation was performed on an extensive dataset comprising anonymized information sourced from six major banks. The algorithms can utilize integrated information encompassing consumer tradelines, credit bureau data, and macroeconomic data spanning from January 2009 through December 2013. Their findings revealed that, in both samples and time frames, decision trees and random forests outperformed logistic regression in predicting credit card risks. This highlights the potential advantages of using big data and machine learning techniques for the benefit of consumers, risk managers, stakeholders, and anyone seeks to avoid unexpected losses and reduce the costs associated with consumer credit.

Sun and Vasarhelyi \unskip~\cite{2069413:29032128} implemented a deep neural network for credit card risk prediction, employing a dataset comprising 711,397 credit card holders from a prominent Brazilian bank. Their study demonstrated that the deep neural network outperformed several other machine learning models, including logistic regression, naive Bayes, traditional artificial neural networks, and decision trees. It achieved the highest F scores and area under the receiver operating characteristic curve (ROC-AUC) among the models tested. Kumar \textit{et al.}\unskip~\cite{2069413:29032818} introduced deep learning with k-means algorithm for credit card scoring prediction, using a Home Credit Default Risk dataset publicly available on Kaggle \unskip~\cite{2069413:29032870}. Their approach invlolved the following steps, including data preprocessing, feature selection, training a deep learning model, and incorporating a decision support system to enhance the accuracy of the deep learning predictions. Their findings indicated that the proposed model delivered good performance, achieving an 87\% accuracy rate when tested on the dataset\unskip~\cite{2069413:29164641}.

Ala'raj \textit{et al.}\unskip~\cite{2069413:29034506} proposed machine learning approach for predicting the consumer behavior, assisting banks management in credit card scoring clients. Their approach consists of three phases: evaluating the probability of single and consecutive payment delinquencies among credit card customers, analyzing customer purchasing behavior, and grouping customers based on their expected loss in a mathematical context. Their implementation consists of two models: the first model assesses the probability of missed payment in the following month for each customer, and the second model focuses on estimating total monthly purchases. The customer behavior grouping is generated from both models, and both models are trained on real credit card transactional datasets. Their experimental findings revealed that their neural network-based model significantly enhanced consumer credit scoring in comparison to traditional machine learning algorithms.  

 Zhu et al. \unskip~\cite{2069413:29035440} applied machine learning models to predict and analyze loan defaults. They conducted a performance comparison among logistic regression, decision tree, XGBoost, and LightGBM using a large dataset from a Chinese bank. Different feature selection techniques were employed to reduce the number of features, including deletion, principal component analysis, feature interaction analysis, and the population stability index. Their findings indicated that LightGBM outperformed the other models in the comparison. Additionally, they identified several factors, such as loan term, loan grade, credit rating, and loan amount, that significantly influenced the predictive outcomes.  Furthermore, Alam \textit{et al}. \unskip~\cite{2069413:29036530} investigated the prediction of default credit card outcomes in imbalanced datasets. In this investigation, they employed various undersampling and oversampling techniques, in addition to utilizing several machine learning models. Their findings indicate that undersampling techniques tend to yield higher accuracy compared to oversampling methods. Moreover, the performance of different classifiers improved significantly when tested on balanced datasets.

This study introduces a new credit card default dataset for an American bank and investigates various machine learning models to enhance the prediction of defaulting credit cards. We first analyze the features of the data, extract the important features, and then select the most relevant ones. Lastly, we compare the prediction performance using the following machine learning models: logistic regression, decision tree, random forest, XGBoost, LightGBM, and neural network.  Therefore, The main objective of this study is:

"To determine how to extract the most important features from the proposed dataset and identify the best-performing machine learning model."

The rest of the paper is structured as follows: Section 2 provides a review of the methodology and techniques employed in this study. Section 3 presents the dataset description and preparation. Section 4 outlines the results obtained from the implementation stage. Finally, Section 5 presents the conclusions and future works.

\section{Methodology}
The following subsections describe the machine learning models used in this study.

\subsection{Logistic regression model}Logistic regression is a statistical and machine learning model used primarily for binary classification tasks \unskip~\cite{2069413:29045701,2069413:29045702}. This model is typically employed when the class label in the problem is categorical and has only two possible outputs, commonly referred to as '0' and '1,' 'yes' and 'no,' or 'positive' and 'negative.' The primary objective of logistic regression is to predict the probability of the class label for a given set of features (input). Logistic regression maps input variables from linear regression to a probability score between 0 and 1 using the sigmoid function, as shown in Equation~(\ref{dfg-eeda27e72c1c}) Equation 1.
\let\saveeqnno\theequation
\let\savefrac\frac
\def\dispfrac{\displaystyle\savefrac}
\begin{eqnarray}
\let\frac\dispfrac
\gdef\theequation{1}
\let\theHequation\theequation
\label{dfg-eeda27e72c1c}
\begin{array}{@{}l}f(z)\;=\;\frac1{1\;+\:e^{-z}}\end{array}
\end{eqnarray}
\global\let\theequation\saveeqnno
\addtocounter{equation}{-1}\ignorespaces 
Where:
\let\saveeqnno\theequation
\let\savefrac\frac
\def\dispfrac{\displaystyle\savefrac}
\begin{eqnarray}
\let\frac\dispfrac
\gdef\theequation{2}
\let\theHequation\theequation
\label{dfg-e5f5fb980ce2}
\begin{array}{@{}l}\;\;z\;=\;\beta_0+\sum_{k=1}^{m}\beta_kx_k\end{array}
\end{eqnarray}
\global\let\theequation\saveeqnno
\addtocounter{equation}{-1}\ignorespaces 
Here, $\beta_0 $ indicates the intercept of the algorithm, and $\beta_1,\;\beta_2,\;...,\;\beta_k\; $, are the coefficients of the independent variables $(x_1,\;x_2,\;...,\;x_k) $ , and \textit{m} is the number of independent variables (number of features).

Logistic regression is widely used in many fields, including medicine\unskip~\cite{2069413:29045845}, finance\unskip~\cite{2069413:29045846}, marketing\unskip~\cite{2069413:29045848,2069413:29164738}, and many other areas where binary classification is needed. Therefore, it is worth implementing and investigating in our credit card scoring system.

\subsection{Decision tree}A decision tree is a machine learning algorithm used for both classification and regression problems. It is a graphical representation of the decision-making process that forms a complete tree structure. Each internal node in the decision tree represents a decision or a test on an input variable, and the outcomes of these tests are represented by branches. Each leaf node, on the other hand, represents a class label of the problem domain\unskip~\cite{2069413:29046266,2069413:29164705}. 

The advantages of using decision trees include their ease of understanding and interpretation, their minimal data preparation requirements, and their low computational cost, which scales logarithmically with the number of data samples used for training. However, a notable disadvantage of decision trees is their sensitivity to overfitting the training data if not properly pruned\unskip~\cite{2069413:29046268}.

\subsection{Random forest}Random Forest is a widely used machine learning algorithm and commonly used for both classification and regression tasks. It extends the concept of decision trees by constructing multiple trees on different subsets of the data and aggregating their predictions through majority voting for classification tasks. One of its primary advantages lies in its capability to handle complex datasets effectively while handling the risk of overfitting. This attribute makes it a robust model known for its high predictive accuracy\unskip~\cite{2069413:29047192}.

\subsection{XGBoost (Extreme Gradient Boosting)}XGBoost is a powerful and high-speed machine learning algorithm that is considered one of the gradient boosting algorithms. It builds a series of decision trees sequentially, with each tree correcting the prediction errors of the previous one\unskip~\cite{2069413:29069368,2069413:29069369}. These generated models are then combined to compute the final output. XGBoost offers several advantages, including the inclusion of L1 and L2 regularization terms in its objective function, making it a robust model that is less prone to overfitting (see Equation~(\ref{dfg-50b95bbdc237})). Additionally, XGBoost employs a technique called 'tree pruning' to reduce the complexity of individual trees. Moreover, the algorithm can handle missing values by learning the best imputation values during training. These advantages collectively make XGBoost an efficient and versatile method\unskip~\cite{2069413:29069371}.
\let\saveeqnno\theequation
\let\savefrac\frac
\def\dispfrac{\displaystyle\savefrac}
\begin{eqnarray}
\let\frac\dispfrac
\gdef\theequation{3}
\let\theHequation\theequation
\label{dfg-50b95bbdc237}
\begin{array}{@{}l}obj^{(t)}\;=\;\sum_{i=1}^{n}l(y_i,\overset\frown{y_l}^{t}\;)\;+\:\sum_{i=1}^{t}\Omega(f_t)\end{array}
\end{eqnarray}
\global\let\theequation\saveeqnno
\addtocounter{equation}{-1}\ignorespaces 
Here $l(y_i,\;\overset\frown{y_l}^{t}) $ is the general loss function of the t\ensuremath{^{th}}  boosted tree and and $\Omega(f_t) $ is the leguraization term that used to reduce teh overfitting as in ~ : 
\let\saveeqnno\theequation
\let\savefrac\frac
\def\dispfrac{\displaystyle\savefrac}
\begin{eqnarray}
\let\frac\dispfrac
\gdef\theequation{4}
\let\theHequation\theequation
\label{dfg-43cf1400015c}
\begin{array}{@{}l}\Omega(f)=\frac12\lambda\sum_{j=1}^{J}w_j^{2}\end{array}
\end{eqnarray}
\global\let\theequation\saveeqnno
\addtocounter{equation}{-1}\ignorespaces 
Where $w $ represents the weights of the leaves  and $\lambda $ is the regularization parameter that control the overfitting.

\subsection{LightGBM (Light Gradient Boosting Machine) }LightGBM is an efficient, high-speed, and scalable machine learning model that belongs to the gradient boosting family. It sequentially constructs an ensemble of decision trees to enhance predictive accuracy. LightGBM utilizes gradient-based one-sided sampling (GOSS) to expedite the training process. Another technique it employs is exclusive feature bundling (EFB), which can combine exclusive features that take nonzero values simultaneously. EFB serves as a dimensionality reduction method by merging certain features.

\subsection{Multi layer perceptron neural network (MLP)}MLP is an artificial neural network composed of multiple layers of connected nodes. It often referred as a feedforward neural network because data flows in one direction, from input layer through the hidden layer, and finally reaching the output layer. The MLP consists of three main layers: input layer, one or more hidden layer, and output layer. The process begins with the input layer receiving the raw data, which is then processed as it goes through the hidden layers. Each layer contains neurons that perform operations involving the weighted sum of their inputs, the application of an activation function (commonly ReLU), and the passing of the output to the next layer (see Figure~\ref{f-a6b9b7fde70d}). The network's prediction is usually computed in the output layer, providing the final result \unskip~\cite{2069413:29071945,2069413:29071946,2069413:29071947}. More details about MLP neural network can be found in the study by Ramchoun \textit{et al}. \unskip~\cite{2069413:29071971,2069413:29164643}.
\bgroup
\fixFloatSize{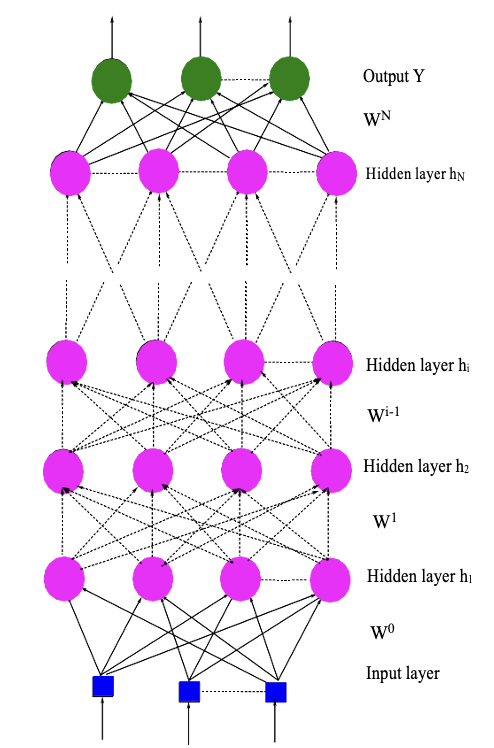}
\begin{figure}[!htbp]
\centering \makeatletter\IfFileExists{images/1a168be8-8efb-4b4d-bf6f-9838f7824769-ufig-1.png}{\includegraphics[width=.73\linewidth]{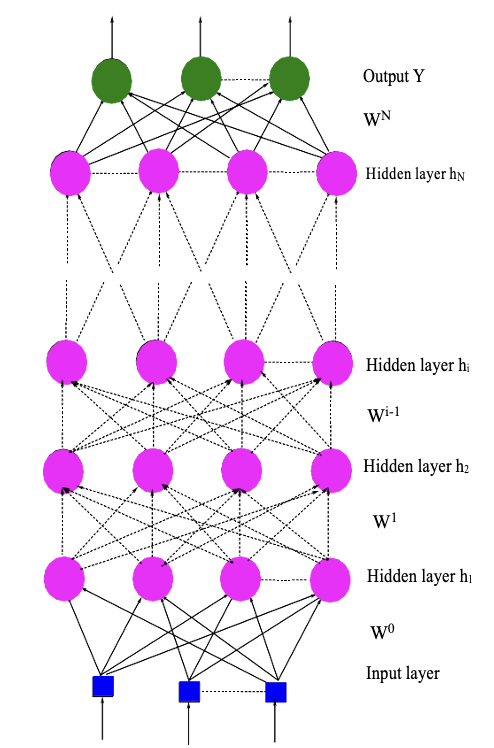}}{\includegraphics{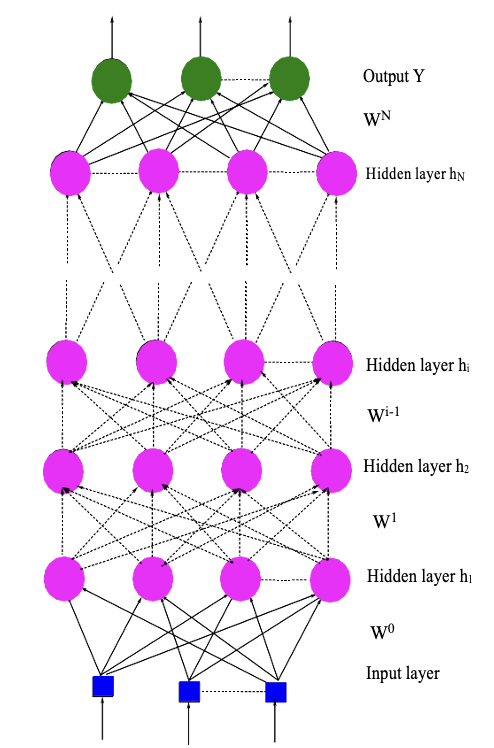}}
\makeatother 
\caption{{Feed forward neural network structure \unskip~\protect\cite{2069413:29071971}}}
\label{f-a6b9b7fde70d}
\end{figure}
\egroup

\section{Data preparation }
The following subsections describe the data source and the preprocessing steps that have been done to this dataset.

\subsection{Data sources}The data used in this study is a new dataset collected from an American bank, and it is available upon request. The dataset was collected over the last 12 months from the data collection date, and the features of the data are listed in Table~\ref{tw-25e3570abf34}. The dataset consists of 500 entries, each containing 36 feature. Twelve of these features are described, while the remaining 24 are anonymized and labeled as var0 to var12. In this dataset, 'Target' indicates the default state, where '1' represents default customers and '0' represents non-default customers. Table~\ref{tw-3bd00bd3def1} provides a statistical results for the dataset.
\begin{table}[!htbp]
\caption{{Dataset features description} }
\label{tw-25e3570abf34}
\def\arraystretch{1}
\ignorespaces 
\centering 
\begin{tabulary}{\linewidth}{LLL}
\hline No. & Feature & Description\\
\hline 
1 &
  Payroll\_1 &
  Payroll of month 1\\
2 &
  Payroll\_2 &
  Payroll of month 1\\
... &
  ... &
  ...\\
12 &
  Payroll\_12 &
  Payroll of month 12\\
13 &
  Var0 &
  Anonymized feature\\
14 &
  Var1 &
  Anonymized feature\\
... &
  ... &
  ...\\
36 &
  Var24 &
  Anonymized feature\\
\hline 
\end{tabulary}\par 
\end{table}

\begin{table*}[!htbp]
\caption{{Statistical results of the dataset} }
\label{tw-3bd00bd3def1}
\def\arraystretch{1}
\ignorespaces 
\centering 
\begin{tabulary}{\linewidth}{LLLLLLL}
\hline  & payroll\_1 & payroll\_6 & payroll\_12 & Var0 & Var12 & Var24\\
\hline 
Count &
  477.0 &
  477.0 &
  477.0 &
  477.0 &
  477.0 &
  477.0\\
Mean &
  928.788 &
  745.71 &
  487.91 &
  0.12 &
  0.01 &
  0.14\\
Std. &
  1274.89 &
  1216.0 &
  1041.09 &
  2.08 &
  1.82 &
  0.38\\
Min &
  0.0 &
  0.0 &
  0.0 &
  -33.37 &
  -37.82 &
  0.0\\
25\% &
  0.0 &
  0.0 &
  0.0 &
  0.0 &
  0.0 &
  0.0\\
50\% &
  508.3 &
  0.0 &
  0.0 &
  0.0 &
  0.0 &
  0.0\\
75\% &
  1221.38 &
  1000.13 &
  465.09 &
  0.17 &
  0.0 &
  0.0\\
Max &
  7487.63 &
  6750.12 &
  5996.12 &
  17.45 &
  6.06 &
  3.30\\
\hline 
\end{tabulary}\par 
\end{table*}

\subsection{Data preprocessing}In the following subsection, we will present and discuss the three main preprocessing steps.

\subsubsection{Handling missing values}To prepare the data for ML models, we applied several preprocessing techniques to our dataset. Firstly, we handled missing values by assigning a 0 value to all missing entries. Since the first 12 variables represent monthly payrolls, any missing value in these variables is considered as the customer not having a payroll for that month. Consequently, for the remaining 24 variables, which are extracted based on the first 12, we assigned 0 values to any missing entries.

\subsubsection{Handling outliers} This study also addresses the issue of outliers to ensure that they do not affect the results of ML models. Our method employs standard deviation and calculates the upper boundary and lower boundary by adding and subtracting 3 standard deviations from the mean of the values. According to this method, outliers are defined as values that fall beyond the range of (\textit{\ensuremath{\mu }}-3\ensuremath{\sigma  }, \textit{\ensuremath{\mu }}+3\ensuremath{\sigma  }), where \textit{\ensuremath{\mu } }represents the mean value, and \ensuremath{\sigma  } represents the standard deviation. When the data follows a normal distribution, the probability of data values falling outside of this range is less than 0.3\%. Next, we applied the Winsorization process by replacing the extreme values. For example, values exceeding the upper boundary were replaced with the upper boundary value, while values below the lower boundary were replaced with the lower boundary value.

\subsubsection{Handling imbalanced dataset}Our new dataset exhibits class imbalance, where one class has a larger number of observations, known as the majority class, while the other class has a smaller number of observations, known as the minority class. This class imbalance issue can negatively affect the performance of machine learning models. In our dataset, the number of default customers is larger than the number of non-default customers, primarily because the bank receives a large number of loan applications, and a significant portion of these applications is rejected for various reasons. The distribution of classes in our dataset is depicted in Figure~\ref{f-404a18f48085}.

\bgroup
\fixFloatSize{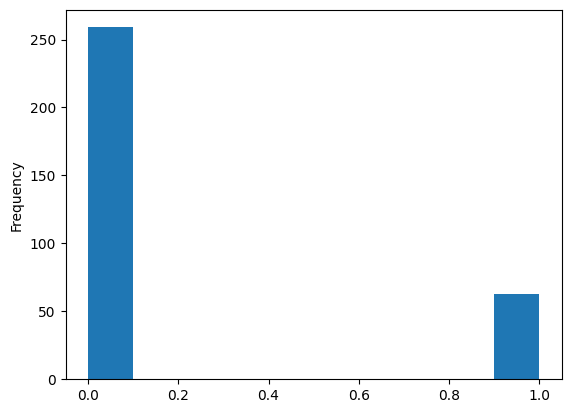}
\begin{figure}[!htbp]
\centering \makeatletter\IfFileExists{images/66611410-b8da-45c8-970b-150253137255-uimbalanced.png}{\includegraphics[width=.91\linewidth]{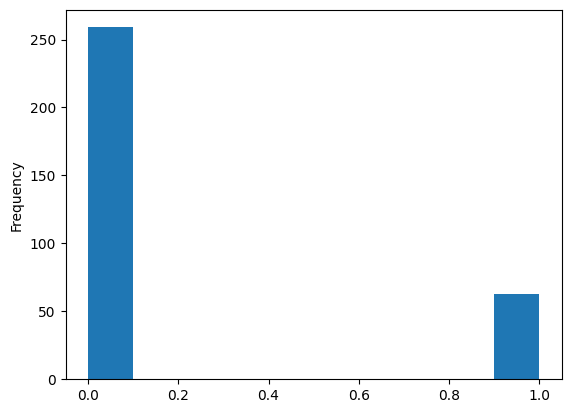}}{\includegraphics{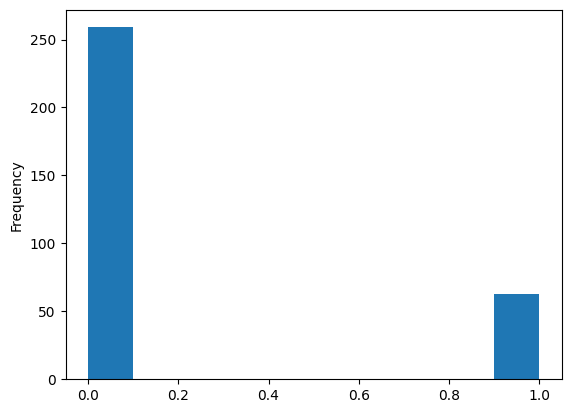}}
\makeatother 
\caption{{The distribution of class labels}}
\label{f-404a18f48085}
\end{figure}
\egroup
To address this issue, we have implemented five different oversampling balancing techniques, namely SMOTE, KMeansSMOTE, BorderlineSMOTE, SVMSMOTE, and RandomOverSampler \unskip~\cite{2069413:29087644}. In preliminary experiments, KMeansSMOTE performed the best and, therefore, it has been selected as a balancing technique for the proposed dataset.
    
\section{Evaluation metrics}
Evaluation metrics are important measures to assess the performance of machine learning models. These metrics also play a crucial role in comparing ML models against each other and discussing how well each model is performing. All evaluation metrics are calculated based on four type of classifications: true positive (TP), false positive (FP), true negative (TN), and false negative (FN). The explanation of these four as follows:

\begin{itemize}
  \item \relax TP: This represents positive data that has been correctly classified as positive.
  \item \relax FP: These are negative data points that have been incorrectly classified as positive.
  \item \relax TN: This category encompasses negative data points that have been correctly classified as negative.
  \item \relax FN: FN includes positive data points that have been incorrectly classified as negative.
\end{itemize}
  The following four subsections explain the four evaluation matrics that are used in this study \unskip~\cite{2069413:29035440,2069413:29036530}.

\subsection{Accuracy}Accuracy is the most widely used classification performance metric. It calculates the number of correctly classified samples by the model out of the total samples in the dataset. Accuracy is calculated as explained in Equation~(\ref{dfg-bb785b440983})
\let\saveeqnno\theequation
\let\savefrac\frac
\def\dispfrac{\displaystyle\savefrac}
\begin{eqnarray}
\let\frac\dispfrac
\gdef\theequation{5}
\let\theHequation\theequation
\label{dfg-bb785b440983}
\begin{array}{@{}l}\text{Accuracy}=\frac{\text{TP + TN}}{\text{TP + FP + TN + FN}}\end{array}
\end{eqnarray}
\global\let\theequation\saveeqnno
\addtocounter{equation}{-1}\ignorespaces

\subsection{Precision}Precision is a metric used to evaluate the accuracy of correctly classified positive samples by the model. It is calculated by dividing the number of TP by the sum of TP and FP. In simpler terms, precision measures how many of the classified TP instances are actually positive. Equation~(\ref{dfg-a2aa95af4e47}) explains how precision is calculated.
\let\saveeqnno\theequation
\let\savefrac\frac
\def\dispfrac{\displaystyle\savefrac}
\begin{eqnarray}
\let\frac\dispfrac
\gdef\theequation{6}
\let\theHequation\theequation
\label{dfg-a2aa95af4e47}
\begin{array}{@{}l}\text{Precision} = \frac{\text{TP}}{\text{TP + FP}}
\end{array}
\end{eqnarray}
\global\let\theequation\saveeqnno
\addtocounter{equation}{-1}\ignorespaces

\subsection{Recall}Recall is a metric used to evaluate the ability of the model in classifying all positive samples correctly. It is also known as sensitivity, or the true positive rate. Recall is calculated by dividing the number of TP by the sum of TP and FN. The equation for recall is explained below in Equation~(\ref{dfg-f447b2c1487a}) :
\let\saveeqnno\theequation
\let\savefrac\frac
\def\dispfrac{\displaystyle\savefrac}
\begin{eqnarray}
\let\frac\dispfrac
\gdef\theequation{7}
\let\theHequation\theequation
\label{dfg-f447b2c1487a}
\begin{array}{@{}l}\text{Recall}=\frac{\text{TP}}{\text{TP +~FN}}\end{array}
\end{eqnarray}
\global\let\theequation\saveeqnno
\addtocounter{equation}{-1}\ignorespaces 
In simpler terms, recall measures how well the model identifies all positive examples in the dataset. A high recall value indicates that the model is effective at minimizing false negatives and is successful in correctly identifying most of the positive cases.

\subsection{Area Under the ROC Curve (AUC-ROC)}Area Under the ROC Curve (AUC-ROC) is an evaluation metric used to measure the area under the Receiver Operating Characteristic (ROC) curve. This curve plots the True Positive Rate (Sensitivity) against the False Positive Rate (1 {\textemdash} Specificity) at various threshold values. The area under the ROC curve, often referred to as the AUC, quantifies the model's ability to distinguish between positive and negative samples. The AUC value falls within the range of 0.5 to 1. A perfect classifier is represented by an AUC value of 1, while an AUC of 0.5 indicates that the model performs no better than random guessing. In general, a higher AUC-ROC score indicates a better-performing model, and it is often used to compare different models' performance on the same dataset. 
    
\section{Experimental results and discussion }
In this section, a comparative analysis is conducted between the proposed ML models to investigate the best performing one for credit card scoring. The dataset used here is the newly proposed dataset explained in Section 3.1. 80\% of the dataset has been randomly selected for the training set, and the remaining 20\% reserved for testing.  In addition, an ANOVA statistical test is conducted to determine whether the differences between the models are statistically significant or not. 

In this discussion, we will analyze the results as presented in Table~\ref{tw-aa3730bcf59e}, which includes accuracy, AUC (Area Under the ROC Curve), precision, and recall for each of the models tested. From the table we can have the following observations:

\textbf{Logistic Regression:} This model achieved reasonable accuracy of 86\%, but it suffers in terms of precision and recall. The precision is 50\%, which indicates that only half of the positive predictions are true positive. The recall is 40\%, which suggests that the model struggles to correctly classify all positive instances. This could be due to the simplicity of the model, as Logistic Regression assumes linear relationships between features.

\textbf{XGBoost Classifier} showed a significant improvement over Logistic Regression. The model performs well in terms of overall correct predictions with an accuracy of 92\%. The AUC of 76\% indicates good discrimination capability. Moreover, it achieves a higher precision of 0.79, indicating better performance in identifying true positives. However, recall value still low (55\%) and there is room for improvement, as it suggests that some positive instances are still missed.

\textbf{LightGBM Classifier} achieved the highest accuracy of 94\%, which indicates a strong ability to make correct predictions. The AUC score of 84\% suggested excellent discrimination between classes. Additionally, the model performed well in precision (82\%) and recall (70\%), striking a good balance between identifying true positives and minimizing false negatives. This model appears to be the best choice for credit card scoring.

\textbf{The Decision Tree Classifier} achieved an accuracy of 86\%, similar to Logistic Regression. However, it demonstrated slightly better AUC (71\%). Precision and recall were both at 50\%, indicating that this model performed moderately but doesn't show good results in distinguishing between classes.

\textbf{Random Forest Classifier} achieved a high accuracy of 93\%, which indicates strong predictive power. However, it's important to note that precision is 1.0, indicating that it classifies all positive predictions correctly. Still, this comes at the cost of recall (50\%), as it fails to capture all positive instances, leading to a trade-off between precision and recall.

\textbf{The MLP Classifier }achieved a good balance between accuracy (92\%) and AUC (83\%). Precision (70\%) and recall (70\%) scores were also balanced, indicating its ability to correctly classify positive instances without overly sacrificing precision. 

Figure~\ref{f-7e394d13e4ae} visualize the AUC graph and represents the performance of each model in distinguishing between positive and negative instances. A higher curve on the graph corresponds to a better-performing model. In the AUC graph, we can see that Random Forest and MLPClassifier show curves that are relatively closer to the upper-left corner, which indicates their superior performance in terms of AUC. Logistic Regression, while having a moderate AUC, lags behind the other models in terms of AUC.

\begin{table}[!htbp]
\caption{{The performance of the proposed ML models} }
\label{tw-aa3730bcf59e}
\def\arraystretch{1}
\ignorespaces 
\centering 
\begin{tabulary}{\linewidth}{LLLLL}
\hline Model & Accuracy (\%) & AUC (\%) & Precision (\%) & Recall (\%)\\
\hline 
LogisticRegression &
  85.42 &
  78.95 &
  48.28 &
  70\\
LightGBM &
  93.75 &
  83.79 &
  82.35 &
  70\\
LightGBM &
  94.44 &
  84.19 &
  87.5 &
  70\\
Decision Tree  &
  89.58 &
  81.37 &
  60.87 &
  70\\
Random Forest  &
  93.06 &
  85.48 &
  75 &
  75\\
MLP Classifier &
  91.67  &
  86.77 &
  66.67 &
  80\\
\hline 
\end{tabulary}\par 
\end{table}
In summary, LightGBM outperformed all compared models in this credit card scoring task, showing high values in accuracy, AUC, precision, and recall. XGBoost and MLP Classifier also performed well and offered a good trade-off between precision and recall. Random Forest demonstrated high accuracy but required further tuning to improve recall. Logistic Regression and Decision Tree Classifier lag behind in performance metrics. Therefore, our observations are twofold: (a) complex models like LightGBM and Random Forest showed superior predictive capabilities compared to traditional logistic regression and decision tree models; (b) LightGBM demonstrated a slightly higher predictive ability than the Random Forest model. However, in this credit scoring dataset, our primary focus is on reducing false negatives to prevent the bank from granting loans to potentially defaulting customers.

Therefore, the Recall evaluation metric, which minimizes false negatives, takes precedence in this study. Based on the presented results, the MLP Classifier considered as the best-performing model for credit card scoring as it has the highest Recall value.
\bgroup
\fixFloatSize{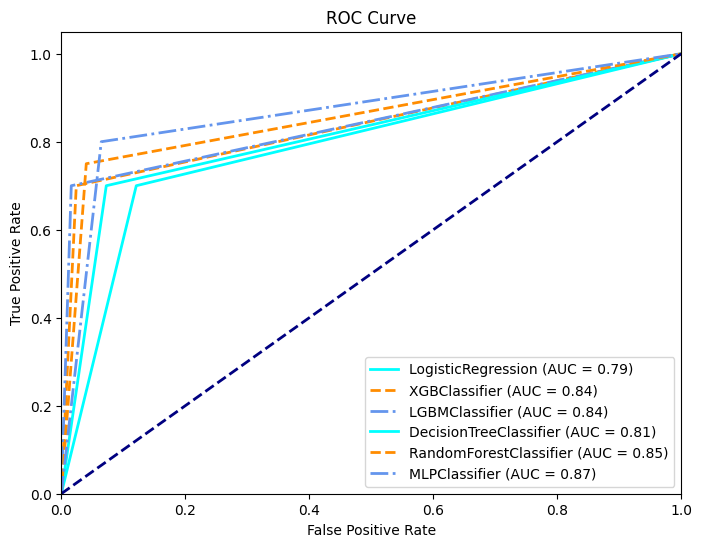}
\begin{figure}[!htbp]
\centering \makeatletter\IfFileExists{images/d69aa12b-1796-4bee-ab7e-1bcd9e30814d-ufig-3.png}{\includegraphics{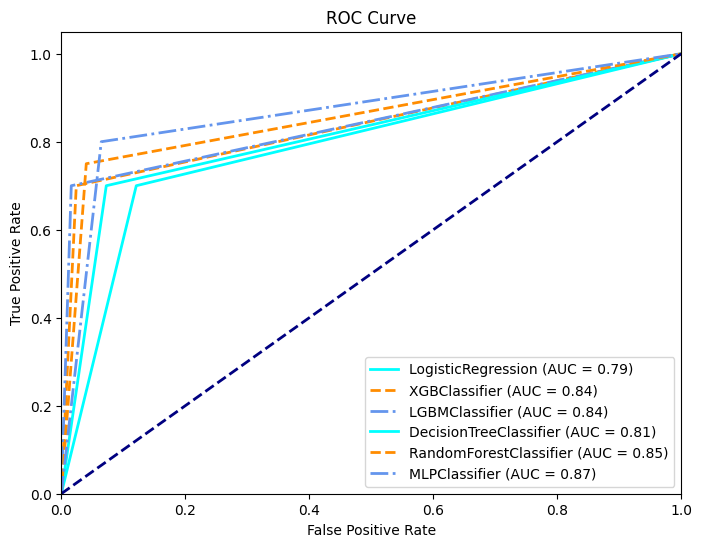}}{\includegraphics{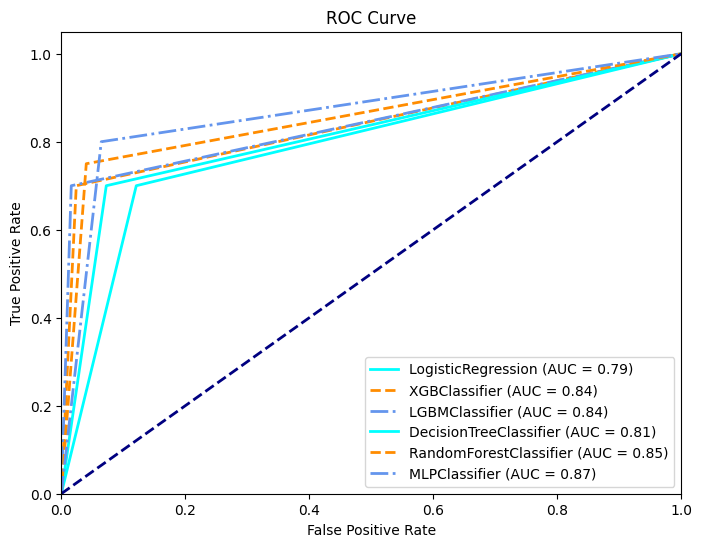}}
\makeatother 
\caption{{ROC curve of all tested models}}
\label{f-7e394d13e4ae}
\end{figure}
\egroup

\section{Conclusion}
In this study, we introduced a new dataset for credit card scoring, sourced from an American bank. We took extra care to preprocess this data, dealing with missing vslues, outliers, and imbalance class label issues. Then, we explored various machine learning techniques like logistic regression, XGBoost, LightGBM, decision trees, random forest, and MLP neural networks.

Our experiments showed that LightGBM performs the best in terms of accuracy. However, since our dataset is primarily concerned with reducing false negatives to protect the bank from potential default customers, we relied on Recall as our key metric. Based on the results, MLP emerged as the top choice for credit card scoring, boasting the highest recall value. This means that MLP is the best model for effectively identifying potential credit card defalt customers. In the future, we aim to enhance the MLP model to improve its accuracy and recall values to make it more powerful model for credit card scoring.

\section*{Acknowledgments}The authors would like to thank the Ministry of Higher Education (MoHE) Malaysia, for the funding and support for this project under Transdisciplinary Research Grant Scheme (TRGS/1/2019/UKM/01/4/1).



%

\bibliographystyle{IEEEtran}

\bibliography{\jobname}
\vfill
\end{document}